\def\year{2021}\relax
\documentclass[letterpaper]{article} % DO NOT CHANGE THIS
\pdfoutput=1 %for arxiv
\usepackage[switch]{lineno}
\usepackage{aaai21}  % DO NOT CHANGE THIS
\usepackage{times}  % DO NOT CHANGE THIS
\usepackage{helvet} % DO NOT CHANGE THIS
\usepackage{courier}  % DO NOT CHANGE THIS
\usepackage[hyphens]{url}  % DO NOT CHANGE THIS
\usepackage{graphicx} % DO NOT CHANGE THIS
\usepackage{natbib}  % DO NOT CHANGE THIS AND DO NOT ADD ANY OPTIONS TO IT
\usepackage{caption} % DO NOT CHANGE THIS AND DO NOT ADD ANY OPTIONS TO IT
\usepackage{multirow}
\usepackage[utf8]{inputenc} % allow utf-8 input
\usepackage{url}            % simple URL typesetting
\usepackage{booktabs}       % professional-quality tables
\usepackage{amsfonts}       % blackboard math symbols
\usepackage{nicefrac}       % compact symbols for 1/2, etc.
\usepackage{microtype}      % microtypography
\newsavebox{\measurebox}

\usepackage{bm}
\usepackage{soul}
\usepackage{amsmath}
\usepackage{amsthm}
\usepackage{bbding} % for the \CheckmarkBold symbol
\usepackage{xcolor}
\usepackage[ruled,vlined,shortend, linesnumbered]{algorithm2e} % algorithm
\usepackage{enumitem}
\usepackage{subcaption}
\usepackage{amsfonts}
\usepackage{algpseudocode}

\newtheorem{theorem}{Theorem}
\newcommand{\codeurl}{https://github.com/Stream-AD/ExGAN}

\title{ExGAN: Adversarial Generation of Extreme Samples}

\author{
Siddharth Bhatia\textsuperscript{\rm 1}\thanks{Equal Contribution},
Arjit Jain\textsuperscript{\rm 2}\footnotemark[1],
Bryan Hooi\textsuperscript{\rm 1}\\
}

\affiliations{
    \textsuperscript{\rm 1}National University of Singapore\\
    \textsuperscript{\rm 2}{IIT Bombay}\\
    siddharth@comp.nus.edu.sg, arjit@cse.iitb.ac.in, bhooi@comp.nus.edu.sg
   
}

\begin{document}
% \linenumbers

\maketitle

\begin{figure*}[!t]
\centering
\includegraphics[width=0.93\textwidth]{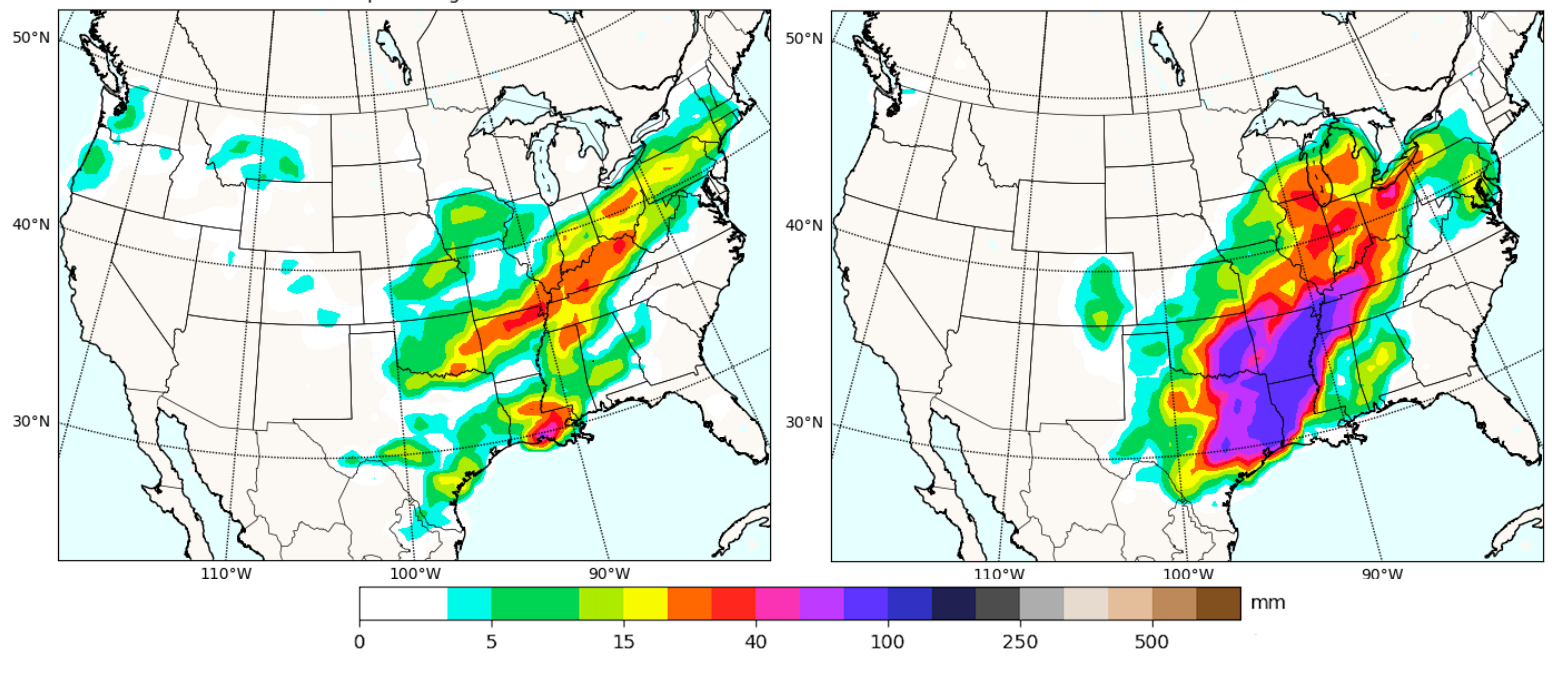}
    \caption{Our goal is to generate samples which are both \emph{realistic} and \emph{extreme}, based on any user-specified extremeness criteria (in this case, high total rainfall). \textit{Left:} Existing GAN-based approaches generate typical rainfall patterns, which have low (green) to moderate (red) rainfall. \textit{Right:} Extreme samples generated by our approach have extreme (violet) rainfall, and realistic spatial patterns resembling that of real floods.}
    \label{fig:teaser}
\end{figure*}

\begin{abstract}
Mitigating the risk arising from extreme events is a fundamental goal with many applications, such as the modelling of natural disasters, financial crashes, epidemics, and many others. To manage this risk, a vital step is to be able to understand or generate a wide range of extreme scenarios. Existing approaches based on Generative Adversarial Networks (GANs) excel at generating realistic samples, but seek to generate \emph{typical} samples, rather than extreme samples. Hence, in this work, we propose ExGAN, a GAN-based approach to generate realistic and extreme samples. To model the extremes of the training distribution in a principled way, our work draws from Extreme Value Theory (EVT), a probabilistic approach for modelling the extreme tails of distributions. For practical utility, our framework allows the user to specify both the desired extremeness measure, as well as the desired extremeness probability they wish to sample at. Experiments on real US precipitation data show that our method generates realistic samples, based on visual inspection and quantitative measures, in an efficient manner. Moreover, generating increasingly extreme examples using ExGAN can be done in constant time (with respect to the extremeness probability $\tau$), as opposed to the $\mathcal{O}(\frac{1}{\tau})$ time required by the baseline approach.
\end{abstract}

\section{Introduction}
Modelling extreme events in order to evaluate and mitigate their risk is a fundamental goal with a wide range of applications, such as extreme weather events, financial crashes, and managing unexpectedly high demand for online services. A vital part of mitigating this risk is to be able to understand or generate a wide range of extreme scenarios. For example, in many applications, stress-testing is an important tool, which typically requires testing a system on a wide range of extreme but realistic scenarios, to ensure that the system can successfully cope with such scenarios. This leads to the question: how can we generate a wide range of extreme but realistic scenarios, for the purpose of understanding or mitigating their risk?

Recently, Generative Adversarial Networks (GANs) and their variants have led to tremendous interest, due to their ability to generate highly realistic samples. On the other hand, existing GAN-based methods generate \emph{typical} samples, i.e. samples that are similar to those drawn from the bulk of the distribution. Our work seeks to address the question: how can we design deep learning-based models which can generate samples that are not just realistic, but also extreme (with respect to any user-specified measure)? Answering this question would allow us to generate extreme samples that can be used by domain experts to assist in their understanding of the nature of extreme events in a given application. Moreover, such extreme samples can be used to perform stress-testing of existing systems, to ensure that the systems remain stable under a wide range of extreme but realistic scenarios. 

Our work relates to the recent surge of interest in making deep learning algorithms reliable even for safety-critical applications such as medical applications, self-driving cars, aircraft control, and many others. Toward this goal, our work explores how deep generative models can be used for understanding and generating the extremes of a distribution, for any user-specified extremeness probability, rather than just generating typical samples as existing GAN-based approaches do. 

More formally, our problem is as follows: Given a data distribution and a criterion to measure extremeness of any sample in this data, can we generate a diverse set of realistic samples with any given extremeness probability? Consider a database management setting with queries arriving over time; users are typically interested in resilience against high query loads, so they could choose to use the number of queries per second as a criterion to measure extremeness. Then using this criterion, we aim to simulate extreme (i.e. rapidly arriving) but realistic query loads for the purpose of stress testing. Another example is rainfall data over a map, as in Figure \ref{fig:teaser}. Here, we are interested in flood resilience, so we can choose to measure extremeness based on total rainfall. Then, generating realistic extreme samples would mean generating rainfall scenarios with spatially realistic patterns that resemble rainfall patterns in actual floods, such as in the right side of Figure \ref{fig:teaser}, which could be used for testing the resilience of a city's flood planning infrastructure.

To model extremeness in a principled way, our approach draws from Extreme Value Theory (EVT), a probabilistic framework designed for modelling the extreme tails of distributions. However, there are two additional aspects to this problem which make it challenging. The first issue is the lack of training examples: in a moderately sized dataset, the rarity of ``extreme" samples means that it is typically infeasible to train a generative model only on these extreme samples. The second issue is that we need to generate extreme samples at any given, user-specified extremeness probability.

One possible approach is to train a GAN, say DCGAN \cite{radford2016unsupervised}, over all the images in the dataset regardless of their extremeness. A rejection sampling strategy can then be applied, where images are generated repeatedly until an example satisfying the desired extremeness probability is found. However, as we show in Section \ref{experiments}, the time taken to generate extreme samples increases rapidly with increasing extremeness, resulting in poor scalability.

Our approach, ExGAN, relies on two key ideas. Firstly, to mitigate the lack of training data in the extreme tails of the data distribution, we use a novel \textbf{distribution shifting} approach, which gradually shifts the data distribution in the direction of increasing extremeness. This allows us to fit a GAN in a robust and stable manner, while fitting the tail of the distribution, rather than its bulk. Secondly, to generate data at any given extremeness probability, we use \textbf{EVT-based conditional generation}: we train a conditional GAN, conditioned on the extremeness statistic. This is combined with EVT analysis, along with keeping track of the amount of distribution shifting performed, to generate new samples at the given extremeness probability.

We present a thorough analysis of our approach, ExGAN, on the US precipitation data. This dataset consists of daily precipitation data over a spatial grid across the lower $48$ United States (Continental United States), Puerto Rico, and Alaska. The criteria used to define extremeness is the total rainfall, and, as explained above, an extreme scenario would correspond to a flood. We show that we are able to generate realistic and extreme rainfall patterns.

\begin{figure*}[t!]
\begin{subfigure}{\textwidth}
\includegraphics[width=\linewidth]{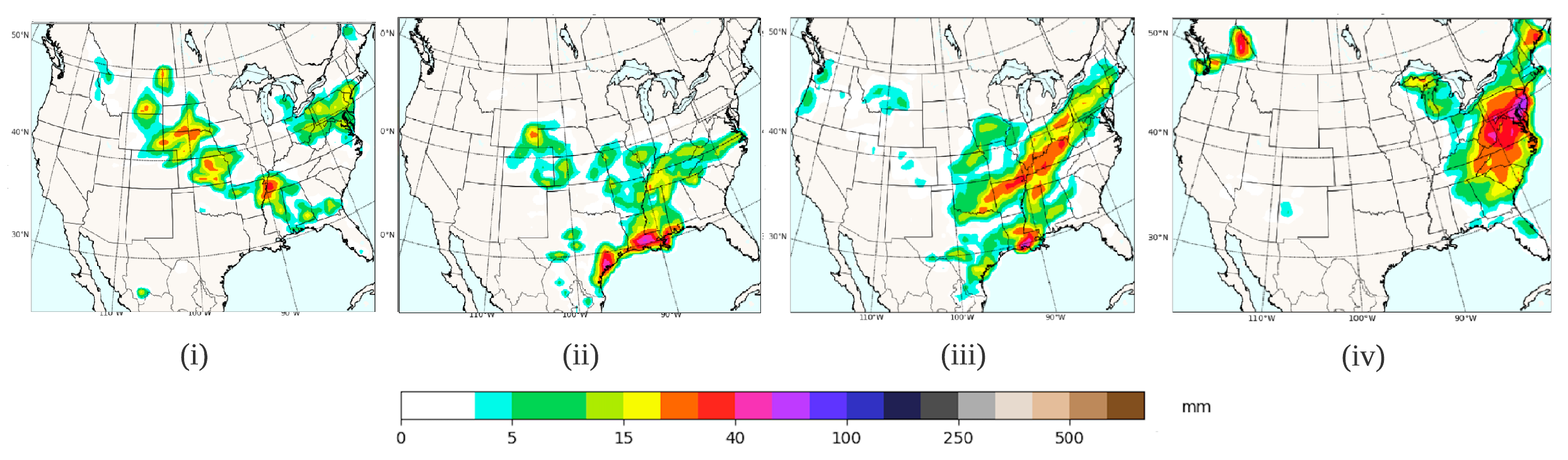}
\caption{Normal samples ((i) and (ii)) from the original dataset show low and moderate rainfall. Samples generated using DCGAN ((iii) and (iv)) are similar to normal samples from the original dataset.} \label{fig:1a}
\end{subfigure}
\begin{subfigure}{\textwidth}
\includegraphics[width=\linewidth]{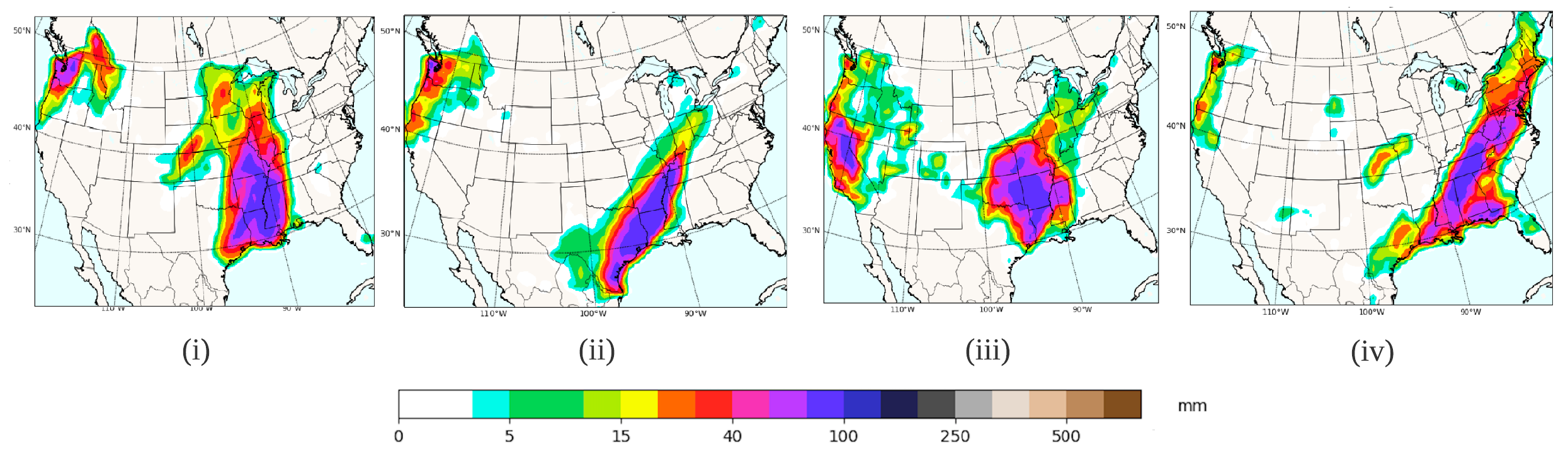}
\caption{Extreme samples ((i) and (ii)) from the original dataset showing high rainfall. Samples generated using ExGAN ((iii) and (iv)) are similar to extreme samples from the original dataset.} \label{fig:1b}
\end{subfigure}
\caption{Comparison between DCGAN (which generates normal samples), and ExGAN (which generates extreme samples).} \label{fig:2}
\end{figure*}

Figure \ref{fig:2} shows images of rainfall patterns from the data, both normal and extreme samples, and images sampled from DCGAN and ExGAN simulating normal and extreme conditions. 

In summary, the main contributions of our approach are:
\begin{enumerate}
    \item {\bf Generating Extreme Samples:} We propose a novel deep learning-based approach for generating extreme data using distribution-shifting and EVT analysis.
    \item {\bf Constant Time Sampling:} We demonstrate how our approach is able to generate extreme samples in constant-time (with respect to the extremeness probability $\tau$), as opposed to the $\mathcal{O}(\frac{1}{\tau})$ time taken by the baseline approach.
    \item {\bf Effectiveness:} Our experimental results show that ExGAN generates realistic samples based on both visual inspection and quantitative metrics, and is faster than the baseline approach by at least three orders of magnitude for extremeness probability of $0.01$ and beyond.
\end{enumerate}

{\bf Reproducibility}: Our code and datasets are publicly available at \codeurl.

\section{Related Work}
\subsection{Conditional Generative Adversarial Networks}
Conditional GANs (CGANs), introduced in \cite{cGAN}, allow additional information as input to GAN which makes it possible to direct the data generation process. Conditional DCGAN (CDCGAN) \cite{cdcgan}, is a modification of CGAN using the conditional variables but with a  convolutional architecture. These methods are discussed briefly in Appendix \ref{app:4}. 
There has also been a significant amount of work done on GAN based models for conditioning on different type of inputs such as images \cite{imagegan1,imagegan2}, text \cite{textgan2}, and multi-modal conditional GANs \cite{multimodal}.

\subsection{Data Augmentation}
Data Augmentation using GANs \cite{antoniou2017data,shmelkov2018good,tran2017bayesian,tran2020towards,yamaguchi2019effective,karras2020training} has been extensively used in different domains, such as anomaly detection \cite{lim2018doping}, time series \cite{zhou2019beatgan,ramponi2018t}, speech processing \cite{zhang2019dada}, NLP \cite{chang2018code,yu2017seqgan, fedus2018maskgan}, emotion classification \cite{zhu2018emotion, luo2018eeg_emotion}, 
medical applications \cite{Zheng2017UnlabeledSG, han2019learning, hu2018prostategan, calimeri2017biomedical} and 
computer vision \cite{karras2019style, odena2017conditional, perez2017effectiveness, sixt2018rendergan, choi2019self, siarohin2019appearance} as a solution for tackling class imbalance \cite{bagan} and generating cross-domain data \cite{crossdomain}. However, these methods do not provide any control over the extremeness of the generated data.

\subsection{Extreme Value Theory}
Extreme value theory~\cite{gumbel2012statistics,theorem2} is a statistical framework for modelling extreme deviations or tails of probability distributions. EVT has been applied to a variety of machine learning tasks including anomaly detection \cite{evtad1, Siffer2017AnomalyDI, evtad3, evtad4, evtad5}, graph mining \cite{hooi2020telltail} and local intrinsic dimensionality estimation \cite{evtlid1, evtlid2}. \cite{evtnips} use EVT to develop a probabilistic framework for classification in extreme regions, \cite{robustnessevt} use it to design an attack-agnostic robustness metric for neural networks.

EVT typically focuses on modelling univariate or low-dimensional~\cite{tawn1990modelling} distributions. A few approaches, such as dimensionality-reduction based~\cite{chautru2015dimension,sabourin2014bayesian}, exist for moderate dimensional vectors (e.g. $20$). A popular approach for multivariate extreme value analysis is Peaks-over-Threshold with specific definitions of exceedances \cite{rootzen2006, ferreira2014, engelke}, and \cite{dombry} showed it can be modelled by r-Pareto processes. \cite{davison2018, fondeville2020functional} presented an inference method on r-Pareto processes applicable to higher dimensions compared to previous works on max-stable processes \cite{asadi} and Pareto processes \cite{thibaud}.

To the best of our knowledge, there has not been any work on extreme sample generation using deep generative models.

\section{Background}  \label{background}

\subsection{Extreme Value Theory (EVT)}

The Generalized Pareto Distribution (GPD) ~\cite{coles2001introduction} is a commonly used distribution in EVT. The parameters of GPD are its \textbf{scale} $\sigma$, and its \textbf{shape} $\xi$. The cumulative distribution function
 (CDF) of the GPD is:
\begin{align}
G_{\sigma, \xi}(x) = 
\begin{cases}
    1-(1+\frac{\xi\cdot x}{\sigma})^{-1/\xi} & \text{if } \xi \neq 0 \\
    1-\exp(-\frac{x}{\sigma}) & \text{if } \xi = 0
\end{cases}
\end{align}

A useful property of the GPD is that it generalizes both Pareto distributions (which have heavy tails) and exponential distributions (which have exponentially decaying tails). In this way, the GPD can model both heavy tails and exponential tails, and smoothly interpolate between them. Another property of the GPD is its `universality' property for tails: intuitively, it can approximate the tails of a large class of distributions following certain smoothness conditions, with error approaching $0$. Thus, the GPD is particularly suitable for modelling the tails of distributions.

\cite{theorem2,theorem1} show that the excess over a sufficiently large threshold $u$, denoted by $X - u$, is likely to follow a Generalized Pareto Distribution (GPD) with parameters $\sigma(u),\xi$. This is also known as the Peaks over Threshold method. In practice, the threshold $u$ is commonly set to a value around the $95^{th}$ percentile, while the remaining parameters can be estimated using maximum likelihood estimation~\cite{grimshaw1993computing}.

\section{ExGAN: Extreme Sample Generation Using GANs}
\label{methods}
\subsection{Problem}
We are given a training set $\mathbf{x}_1, \cdots, \mathbf{x}_n \sim \mathcal{D}$, along with $\mathsf{E}(\mathbf{x})$, a user-defined \emph{extremeness measure}: for example, in our running example of rainfall modelling, the extremeness measure is defined as the total rainfall in $\mathbf{x}$, but any measure could be chosen in general. We are also given a user-specified \emph{extremeness probability} $\tau \in (0, 1)$, representing how extreme the user wants their sampled data to be: for example, $\tau=0.01$ represents generating an event whose extremeness measure is only exceeded $1\%$ of the time.\footnote{In hydrology, the notion of a \emph{$100$-year flood} is a well-known concept used for flood planning and regulation, which is defined as a flood that has a $1$ in $100$ chance of being exceeded in any given year. Given daily data, generating a 100-year flood then corresponds to setting $\tau=\frac{1}{365 \times 100}.$ }

Given these, our goal is to generate synthetic samples $\mathbf{x}'$ that are both 1) \emph{realistic}, i.e. hard to distinguish from the training data, and 2) \emph{extreme} at the given level: that is, 
$P_{\mathbf{x} \sim \mathcal{D}}(\mathsf{E}(\mathbf{x}) > \mathsf{E}(\mathbf{x}'))$ should be as close as possible to $\tau$.

\subsection{Distribution Shifting}
\label{secdistshift}
An immediate issue we face is that we want our trained model to mimic the extreme tails, not the bulk of the distribution; however, most of the data lies in its bulk, with much fewer samples in its tails. While data augmentation could be employed, techniques like image transforms may not be applicable: for example, in the US precipitation data, each pixel captures the rainfall distribution at some fixed location; altering the image using random transforms would change this correspondence.

To address this issue, we propose a novel Distribution Shifting approach in Algorithm \ref{alg:distshift}, parameterized by a shift parameter $c \in (0, 1)$. Our overall approach is to repeatedly `shift' the distribution by filtering away the less extreme $(1-c)$ proportion of the data, then generating data to return the dataset to its original size. In addition, to maintain the desired proportion of original data points from $\mathcal{X}$, we adopt a `stratified' filtering approach, where the original and generated data are filtered separately.

Specifically, we first sort our original dataset $\mathcal{X}$ in decreasing order of extremeness (Line 2), then initialize our shifted dataset $\mathcal{X}_s$ as $\mathcal{X}$ (Line 3). Next, each iteration $i$ of a Distribution Shift operation works as follows. We first fit a DCGAN to $\mathcal{X}_s$ (Line 6). We then replace our shifted dataset $\mathcal{X}_s$ with the top $\lfloor c^i\cdot n\rfloor$ extreme data points from $\mathcal{X}$ (Line 7). Next, we use the DCGAN to generate additional $\lceil (n-\lfloor c^{i}\cdot n\rfloor) \cdot \dfrac{1}{c}\rceil$ data samples and add the most extreme $n-\lfloor c^{i}\cdot n\rfloor$ samples to $\mathcal{X}_s$ (Line 8). This ensures that we choose the most extreme $c$ proportion of the generated data, while bringing the dataset back to its original size of $n$ data points. Each such iteration shifts the distribution toward its upper tail by a factor of $c$. We perform $k$ iterations, aiming to shift the distribution sufficiently so that $\tau$ is no longer in the extreme tail of the resulting shifted distribution. Iteratively shifting the distribution in this way ensures that we always have enough data to train the GAN in a stable manner, while allowing us to gradually approach the tails of the distribution.

 \begin{algorithm}
	\caption{Distribution Shifting\ \label{alg:distshift}}
	{\bf Input}: dataset $\mathcal{X}$, extremeness measure $\mathsf{E}$, shift parameter $c$, iteration count $k$ \\
	Sort $\mathcal{X}$ in decreasing order of extremeness\\
	Initialize $\mathcal{X}_s \gets \mathcal{X}$ \\
    \For{$i \gets 1 \text{ to } k$}{
    {\bf $\triangleright$ Shift the data distribution by a factor of $c$:}\\
    Train DCGAN $G$ and $D$ on $\mathcal{X}_s$\\
	$\mathcal{X}_s \gets$ top $\lfloor c^{i}\cdot n\rfloor$ extreme samples of $\mathcal{X}$ \\
    Generate $\lceil (n-\lfloor c^{i}\cdot n\rfloor) \cdot \dfrac{1}{c}\rceil$ data points using $G$, and insert most extreme $n-\lfloor c^{i}\cdot n\rfloor$ samples into $\mathcal{X}_s$\\
    }
 \end{algorithm}

In addition, during the shifting process, we can train successive iterations of the generator via `warm start', by initializing its parameters using the previous trained model, for the sake of efficiency.

\subsection{EVT-based Conditional Generation}
The next issue we face is the need to generate samples at the user-given extremeness probability of $\tau$. Our approach will be to train a conditional GAN using extremeness as conditioning variable. To generate samples, we then use EVT analysis, along with our knowledge of how much shifting has been performed, to determine the necessary extremeness level we should condition on, to match the desired extremeness probability.

Specifically, first note that after $k$ shifts, the corresponding extremeness probability in the shifted distribution that we need to sample at becomes $\tau' = \tau / c^k$. Thus, it remains to sample from the shifted distribution at the extremeness probability of $\tau'$, which we will do using EVT. Algorithm \ref{alg:condgenEVT} describes our approach: we first compute the extremeness values using $\mathsf{E}$ on each point in $\mathcal{X}_s$: i.e. $e_i = \mathsf{E}(\mathbf{x_i}) \ \forall \ \mathbf{x_i} \in \mathcal{X}_s$ (Line 2). Then we perform EVT Analysis on $e_1, \cdots, e_n$: we fit Generalized Pareto Distribution (GPD) parameters $\sigma, \xi$ using maximum likelihood estimation~\cite{grimshaw1993computing} to $e_1, \cdots, e_n$ (Line 3). Next, we train a conditional DCGAN (Generator $G_s$ and Discriminator $D_s$) on $\mathcal{X}_s$, with the conditional input to $G_s$ (within the training loop of $G_s$) sampled from a GPD with parameters $\sigma, \xi$ (Line 4). In addition to the image, $D_s$ takes in a second input which is $e$ for a generated image $G_s(\mathbf{z}, e)$ and $\mathsf{E}(\mathbf{x})$ for a real image $\mathbf{x}$. An additional loss $\mathcal{L}_{\text{ext}}$ is added to the GAN objective:
\begin{align} 
\mathcal{L}_{\text{ext}} = \mathbb{E}_{\mathbf{z},e}\left[\dfrac{|e-\mathsf{E}({G_s(\mathbf{z}, e)})|}{e}\right]
\end{align}
where $\mathbf{z}$ is sampled from multivariate standard normal distribution and $e$ is sampled from a GPD with parameters $\sigma, \xi$. Note that training using $\mathcal{L}_{\text{ext}}$ requires $\mathsf{E}$ to be differentiable.

$\mathcal{L}_{\text{ext}}$ minimizes the distance between the desired extremeness ($e$) and the extremeness of the generated sample ($\mathsf{E}(G_s(z, e)$). This helps reinforce the conditional generation property and prevents the generation of samples with unrelated extremeness. Using the inverse CDF of the GPD, we determine the extremeness level $e'$ that corresponds to an extremeness probability of $\tau'$:
\begin{align} 
    e' = G^{-1}_{\sigma, \xi}(1-\tau')
    \label{eq:gpd}
\end{align}
where $G^{-1}_{\sigma, \xi}$ is the inverse CDF of the fitted GPD (Line 5). 
Finally, we sample from our conditional DCGAN at the desired extremeness level $e'$ (Line 6). 

\begin{algorithm}
	\caption{EVT-based Conditional Generation \label{alg:condgenEVT}}
	{\bf Input}: shifted dataset $\mathcal{X}_s$, extremeness measure $\mathsf{E}$, adjusted extremeness probability $\tau'$ \\
	Compute extremeness values $e_i = \mathsf{E}(\mathbf{x_i}) \ \forall \ \mathbf{x_i} \in \mathcal{X}_s$ \\
	Fit GPD parameters $\sigma, \xi$ using maximum likelihood~\cite{grimshaw1993computing} on $e_1, \cdots, e_n$ \\
	Train conditional DCGAN ($G_s$ and $D_s$) on $\mathcal{X}_s$ where the conditioning input for $G_s$ is sampled from a GPD with parameters $\sigma, \xi$ \\
	Extract required extremeness level: $e' \gets G^{-1}_{\sigma, \xi}(1-\tau')$\\
	Sample from $G_s$ conditioned on extremeness level $e'$\\
\end{algorithm}

\section{Experiments} \label{experiments}
In this section, we evaluate the performance of ExGAN compared to DCGAN on the US precipitation data. We aim to answer the following questions:

\begin{enumerate}[label=\textbf{Q\arabic*.}]
\item {\bf Realistic Samples (Visual Inspection):} Does ExGAN generate realistic extreme samples, as evaluated by visual inspection of the images?
\item {\bf Realistic Samples (Quantitative Measures):} Does ExGAN generate realistic extreme samples, as evaluated using suitable GAN metrics?
\item {\bf Speed:} How fast does ExGAN generate extreme samples compared to the baseline? Does it scale with high extremeness?
\end{enumerate}

Details about our experimental setup, network architecture and software implementation can be found in Appendix \ref{app:1}, \ref{app:2} and \ref{app:3} respectively.
\paragraph{Dataset:}

We use the US precipitation dataset \footnote{https://water.weather.gov/precip/}. The National Weather Service employs a multi-sensor approach to calculate the observed precipitation with a spatial resolution of roughly $4\times4$ km on an hourly basis. We use the daily spatial rainfall distribution for the duration January 2010 to December 2016 as our training set, and for the duration of January 2017 to August 2020 as our test set. We only retain those samples in our test set which are more extreme, i.e. have higher total rainfall, than the $95^{th}$ percentile in the train set. Images with original size $813\times1051$ are resized to $64\times64$ and normalized between $-1$ and $1$.

\paragraph{Baseline:}

The baseline is a DCGAN \cite{radford2016unsupervised} trained over all the images in the dataset, combined with rejection sampling. Specifically, to generate at a user-specified level $\tau$, we use EVT as in our framework (i.e. Eq. \eqref{eq:gpd}) to compute the extremeness level $e=G^{-1}_{\sigma, \xi}(1-\tau)$ that corresponds to an extremeness probability of $\tau$. We then repeatedly generate images until one is found that satisfies the extremeness criterion within $10\%$ error; that is, we reject the image $\mathbf{x}$ if 
$\left|\dfrac{e - \mathsf{E}(\mathbf{x})}{e}\right| > 0.1$.

\paragraph{Evaluation Metrics:}
We evaluate how effectively the generator is able to mimic the tail of the distribution using FID and Reconstruction Loss metrics. Fréchet Inception Distance (FID) \cite{fid} is a common metric used in the GAN literature to evaluate image samples and has been found to be consistent with human judgement. Intuitively, it compares the distributions of real and generated samples based on their activation distributions in a pre-trained network. However, an ImageNet-pretrained Inception network which is usually used to calculate FID is not suitable for our dataset. Hence, we construct an autoencoder trained on test data, as described above, and use the statistics on its bottleneck activations to compute the FID:
\begin{align*}
\mathrm{FID}=\left\|\bm{\mu_{r}}-\bm{\mu_{g}}\right\|^{2}+\operatorname{Tr}\left(\bm{\Sigma_{r}}+\bm{\Sigma_{g}}-2\left(\bm{\Sigma_{r}} \bm{\Sigma_{g}}\right)^{1 / 2}\right)
\end{align*}
where $\operatorname{Tr}$ denotes the trace of a matrix, $\left(\bm{\mu_{r}}, \bm{\Sigma_{r}}\right)$ and $\left(\bm{\mu_{g}}, \bm{\Sigma_{g}}\right)$ are the mean and covariance of the bottleneck activations for the real and generated samples respectively.

We further evaluate our model on its ability to reconstruct unseen extreme samples by computing a reconstruction loss on the test set \cite{Xiang2017OnTE}.

Letting $\mathbf{\tilde{x}}_1, \cdots, \mathbf{\tilde{x}}_m$ denote the test images, the reconstruction loss for an unconditional generator $G$ is given by, 
$$\mathcal{L}_{\mathrm{rec}}=\frac{1}{m} \sum_{i=1}^{m} \min _{\mathbf{z_i}}\left\|G(\mathbf{z_i})-\mathbf{\tilde{x}}_i\right\|_{2}^{2}$$
where $\mathbf{z_i}$ are the latent space vectors

For an extremeness conditioned generator $G$,
$$\mathcal{L}_{\mathrm{rec\_ext}}=\frac{1}{m} \sum_{i=1}^{m} \min _{\mathbf{z_i}}\left\|G(\mathbf{z_i}, \mathsf{E}(\mathbf{\tilde{x}}_i))-\mathbf{\tilde{x}}_i\right\|_{2}^{2}$$

To compute the reconstruction loss, we initialize the latent space vectors $\mathbf{z_i}$ as the zero vector, and perform gradient descent on it to minimize the objective defined above. We use similar parameters as \cite{Xiang2017OnTE} to calculate the reconstruction loss, i.e. learning rate was set to $0.001$ and number of gradient descent steps was set to $2000$, while we use Adam optimizer instead of RMSprop.

We also evaluate how accurately our method is able to condition on the extremeness of the samples. We use Mean Absolute Percentage Error (MAPE), where the error is calculated between the extremeness used to generate the sample ($e$) and the extremeness of the generated sample ($\mathsf{E}(G(\mathbf{z}, e))$).

\begin{align} 
\text{MAPE} = \mathbb{E}_{\mathbf{z},e}\left[\dfrac{|e-\mathsf{E}({G_s(\mathbf{z}, e)})|}{e}\right] \times 100\%
\end{align}
where $\mathbf{z}$ is sampled from multivariate standard normal distribution and $e$ is sampled from a GPD with parameters $\sigma, \xi$.

\subsection{Realistic Samples (Visual Inspection)}
\begin{figure*}[t!]
\begin{subfigure}{\textwidth}
\includegraphics[width=\linewidth]{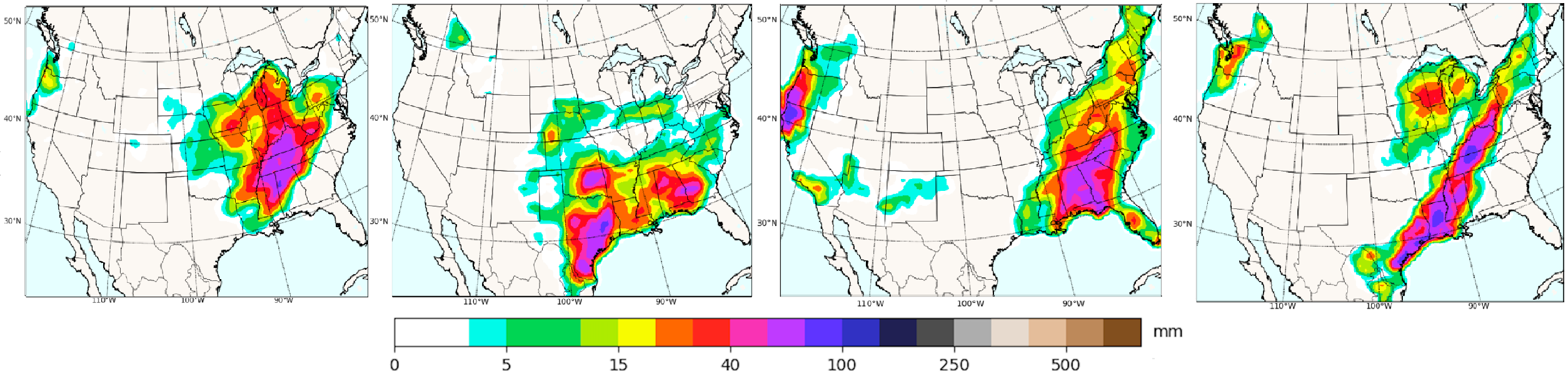}
\caption{Samples from ExGAN for extremeness probability $\tau = 0.001$. Time taken to sample = $0.002s$} \label{fig:3a}
\end{subfigure}
\begin{subfigure}{\textwidth}
\includegraphics[width=\linewidth]{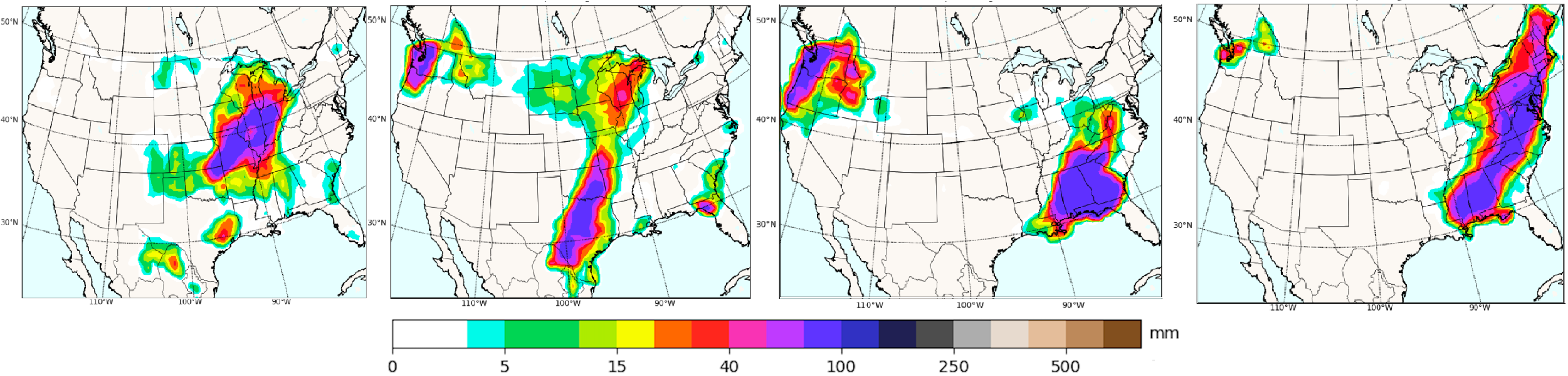}
\caption{Samples from ExGAN for extremeness probability $\tau = 0.0001$. Time taken to sample = $0.002s$} \label{fig:3b}
\end{subfigure}
\begin{subfigure}{\textwidth}
\includegraphics[width=\linewidth]{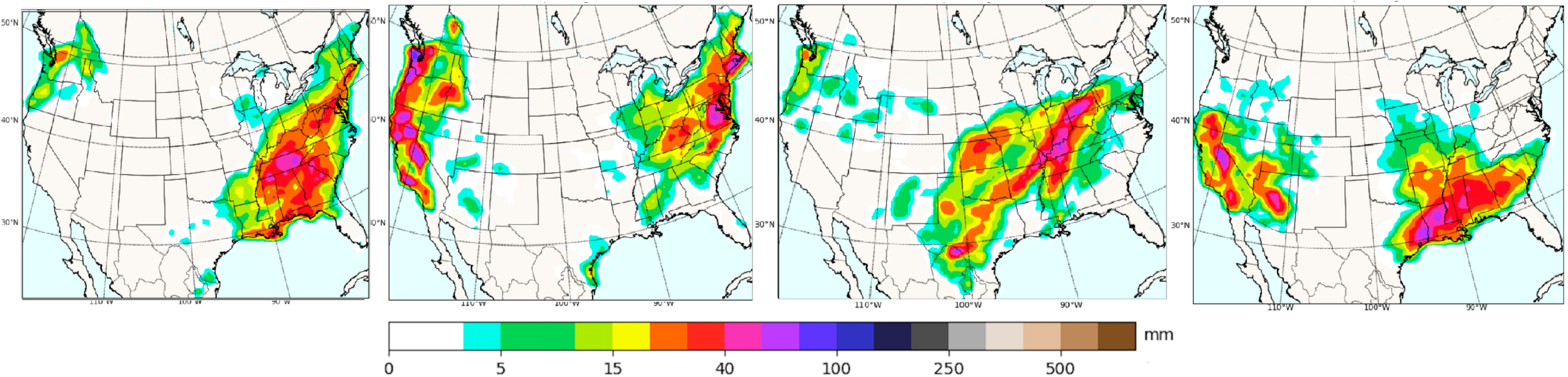}
\caption{Samples from DCGAN for extremeness probability $\tau = 0.01$. Time taken to sample = $7.564s$. DCGAN is unable to generate samples in $1$ hour when $\tau = 0.001$ or $0.0001$.} \label{fig:3c}
\end{subfigure}
\caption{ExGAN generates images which are realistic, similar to the original data samples, in constant time.}
\label{fig:3}
\end{figure*}

Figure \ref{fig:3} shows the extreme samples generated by ExGAN corresponding to extremeness probability $\tau = 0.001$ and $0.0001$. We observe that ExGAN generates samples that are similar to the images of rainfall patterns from the original data in Figure \ref{fig:1b}. As we change $\tau$ from $0.001$ to $0.0001$, we observe the increasing precipitation in the generated samples. The typical pattern of radially decreasing rainfall in real data is learned by ExGAN. ExGAN also learns that coastal areas are more susceptible to heavy rainfall.

Figure \ref{fig:3c} shows the extreme samples generated by DCGAN for extremeness probability $\tau = 0.01$. When $\tau = 0.001$ or $0.0001$, DCGAN is unable to generate even one sample, within $10$\% error, in $1$ hour (as we explain further in Section \ref{sec:speed}).

\subsection{Realistic Samples (Quantitative Measures)}

The GAN is trained for $100$ epochs in each iteration of distribution shifting. For distribution shifting, we set $c=0.75$, $k=10$ and use warm start. MAPE for DCGAN can be upper bounded by the rejection strategy used for sampling, and this bound can be made tighter at the expense of sampling time. For our experiment, we upper bound the MAPE for DCGAN by $10\%$ as explained above. MAPE for ExGAN is $3.14\% \pm 3.08\%$.

Table \ref{tab:metrics} reports the FID (lower is better) and reconstruction loss (lower is better). ExGAN is able to capture the structure and extremeness in the data, and generalizes better to unseen extreme scenarios, as shown by the lower reconstruction loss and lower FID score (loss = $0.0172$ and FID = $0.0236 \pm 0.0037$) as compared to DCGAN (loss = $0.0292$ and FID = $0.0406 \pm 0.0063$).

\begin{table}[ht]
\begin{center}
\caption{FID, and Reconstruction Loss, for DCGAN and ExGAN (averaged over $5$ runs). For FID, the p-value for significant improvement of ExGAN over the baseline is $0.002$, using a standard two-sample t-test.}
\begin{tabular}{ccc}
\toprule

{\bf Method} & {\bf FID} & {\bf Reconstruction Loss} \\
\midrule
\textbf{DCGAN}       & $0.0406 \pm 0.0063$ & $0.0292$ \\
%\textbf{CDCGAN}        & $0.0493 \pm 0.0097$ & $0.0166$\\
\textbf{ExGAN}       & $0.0236 \pm 0.0037$ & $0.0172$ \\

\iffalse
\textbf{ExGAN (c=0.24, k=2)}       & $0.0173$ & $3.43 \pm 3.01$\\
\textbf{ExGAN (c=0.49, k=4)}       & $0.0173$ & $3.32 \pm 3.10$\\
\textbf{ExGAN (c=0.75, k=10)}       & $0.0172$ & $3.14 \pm 3.08$\\
\textbf{ExGAN (c=0.90, k=27)}       & $0.0169$ & $3.05 \pm 3.14$\\
\fi

\bottomrule
\end{tabular}
\label{tab:metrics}
\end{center}
\end{table}

Table \ref{tab:apptable2} reports the reconstruction loss, MAPE and FID for ExGAN for different values of $c$ and $k$. To ensure a fair comparison, we select the parameters $c$ and $k$ for distribution shifting, such that the amount of shift, $c^k$, is approximately similar.
Intuitively, we would expect higher $c$ to correspond to slower and more gradual shifting, which in turn helps the network smoothly interpolate and adapt to the shifted distribution, leading to better performance. This trend is observed in Table \ref{tab:apptable2}. However, these performance gains with higher $c$ values come at the cost of training time.

In Appendix \ref{app:5}, we report ablation results on distribution shifting illustrating its benefit to our approach.

\begin{table}[!ht]
\begin{center}
\caption{Reconstruction Loss, MAPE and FID values for ExGAN for different $c$ and $k$ (averaged over $5$ runs).}
\begin{tabular}{ccccc}
\toprule

$c$&$k$&{\bf Rec. Loss}&{\bf MAPE}&{\bf FID}\\

\midrule

$0.24$ & $2$ & $0.0173$ & $3.43 \pm 3.01$ & $0.0367 \pm 0.0096$\\
$0.49$ & $4$ & $0.0173$ & $3.32 \pm 3.10$ & $0.0304 \pm 0.0109$\\
$0.75$ & $10$ & $0.0172$ & $3.14 \pm 3.08$ & $0.0236 \pm 0.0037$\\
$0.90$ & $27$ & $0.0169$ & $3.05 \pm 3.14$ & $0.0223 \pm 0.0121$\\

\bottomrule
\end{tabular}
\label{tab:apptable2}
\end{center}
\end{table}

\subsection{Speed} \label{sec:speed}

The time taken to generate $100$ samples for different extremeness probabilities is reported in Table \ref{tab:times}. Note that ExGAN is scalable and generates extreme samples in constant time as opposed to the $\mathcal{O}(\frac{1}{\tau})$ time taken by DCGAN to generate samples with extremeness probability $\tau$. DCGAN could not generate even one sample for extremeness probabilities $\tau = 0.001$ and $\tau = 0.0001$ in $1$ hour. Hence, we do not report sampling times on DCGAN for these two values.

\begin{table}[ht!]
\centering
\caption{Sampling times for DCGAN and ExGAN for different extremeness probabilities (in seconds).}
\begin{center}
\begin{tabular}{ccccc}
\toprule
\multirow{2}{*}{\bf Method} & \multicolumn{4}{c}{\bf Extremeness Probability ($\tau)$}\\
  &$0.05$  &$0.01$ &$0.001$ &$0.0001$\\  \midrule
\textbf{DCGAN}    & $1.230s$ & $7.564s$ & $-$ & $-$\\
\textbf{ExGAN}   & $0.002s$ & $0.002s$ & $0.002s$ & $0.002s$\\
\bottomrule
\end{tabular}
\label{tab:times}
\end{center}
\end{table}

\section{Conclusion}
In this paper, we propose ExGAN, a novel deep learning-based approach for generating extreme data. We use (a) distribution shifting to mitigate the lack of training data in the extreme tails of the data distribution; (b) EVT-based conditional generation to generate data at any given extremeness probability.

We demonstrate how our approach is able to generate extreme samples in constant-time (with respect to the extremeness probability $\tau$), as opposed to the $\mathcal{O}(\frac{1}{\tau})$ time taken by the baseline. Our experimental results show that ExGAN generates realistic samples based on both visual inspection and quantitative metrics, and is faster than the baseline approach by at least three orders of magnitude for extremeness probability of $0.01$ and beyond.

The flexibility and realism achieved by the inclusion of GANs, however, comes at the cost of theoretical guarantees. While our algorithmic steps (e.g. Distribution Shifting) are designed to approximate the tails of the original distribution in a principled way, it is difficult to provide guarantees due to its GAN framework. Future work could consider different model families (e.g. Bayesian models), toward the goal of deriving theoretical guarantees, as well as incorporating neural network based function approximators to learn a suitable extremeness measure ($\mathsf{E}$).

\section*{Ethical Impact}
Modelling extreme events in order to evaluate and mitigate their risk is a fundamental goal in a wide range of applications, such as extreme weather events, financial crashes, and managing unexpectedly high demand for online services. Our method aims to generate realistic and extreme samples at any user-specified probability level, for the purpose of planning against extreme scenarios, as well as stress-testing of existing systems. Our work also relates to the goal of designing robust and reliable algorithms for safety-critical applications such as medical applications, aircraft control, and many others, by exploring how we can understand and generate the extremes of a distribution.

Our work explores the use of deep generative models for generating realistic extreme samples, toward the goal of building robust and reliable systems. Possible negative impact can arise if these samples are not truly representative or realistic enough, or do not cover a comprehensive range of possible extreme cases. Hence, more research is needed, such as for ensuring certifiability or verifiability, as well as evaluating the practical reliability of our approach for stress-testing in a wider range of real-world settings.

\bibliography{paper}

\appendix

\section*{Appendix}
\section{Experimental Setup}
\label{app:1}

All experiments are carried out on a $2.6 GHz$  Intel Xeon CPU, $256 GB$ RAM, $12 GB$ Nvidia GeForce RTX 2080 Ti GPU running Debian GNU/Linux $9$. The network architecture and implementation details can be found in Appendix \ref{app:2} and \ref{app:3} respectively.

Images are upsampled from $64\times64$ to $813\times1051$ to plot the rainfall maps. We also apply techniques introduced in the literature to stabilize GAN training such as label smoothing, noisy inputs to the discriminator, lower learning rate for the discriminator, label flipping and gradient clipping~\cite{wgan,noisy}. Details of these techniques can be found in Appendix \ref{app:3}.

\section{Network Architectures}
\label{app:2}

Let ConvBlock denote the sequence of layers Conv$4\times4$, InstanceNorm\cite{instancenorm}, LeakyReLU with appropriate sizes. Similarly let ConvTBlock denote the sequence of layers ConvTranspose4x4, InstanceNorm, LeakyRelu with appropriate sizes. Let $n$ be the batch size.

\begin{table}[!ht]
\centering
\caption{Architecture for ExGAN Generator.}
\label{tab:exgang}
\begin{center}
\begin{tabular}{@{}rccc@{}}
\toprule
 \textbf{Index} & \textbf{Layer} & \textbf{Output Size}  \\ \midrule
$1$ \ \ \ \ & ConvTBlock & $n\times512\times4\times4$\\
$2$ \ \ \ \ & ConvTBlock & $n\times256\times8\times8$\\
$3$ \ \ \ \ & ConvTBlock & $n\times128\times16\times16$\\
$4$ \ \ \ \ & ConvTBlock & $n\times64\times32\times32$\\
$5$ \ \ \ \ & ConvTranpose$4\times4$ & $n\times1\times64\times64$\\
$6$ \ \ \ \ & Tanh & $n\times1\times64\times64$\\
\bottomrule
\end{tabular}

\end{center}
\end{table}
% \FloatBarrier

% \FloatBarrier
\begin{table}[htb!]
\caption{Architecture for ExGAN Discriminator.}
\label{tab:exgand}
\begin{center}
\begin{tabular}{@{}rccc@{}}
\toprule
 \textbf{Index} & \textbf{Layer} & \textbf{Output Size}  \\ \midrule
$1$ \ \ \ \ & ConvBlock & $n\times64\times32\times32$\\
$2$ \ \ \ \ & ConvBlock & $n\times128\times16\times16$\\
$3$ \ \ \ \ & ConvBlock & $n\times256\times8\times8$\\
$4$ \ \ \ \ & ConvBlock & $n\times512\times4\times4$\\
$5$ \ \ \ \ & Conv$4\times4$ & $n\times64\times1\times1$\\
$6$ \ \ \ \ &Reshape &$n\times64$\\
$7$ \ \ \ \ &Concat &$n\times65$\\
$8$ \ \ \ \ & Linear &$n\times1$\\
$9$ \ \ \ \ & Sigmoid &$n\times1$\\
\bottomrule
\end{tabular}
\end{center}
\end{table}

\begin{table}[!ht]
\caption{Architecture for DCGAN Generator.}
\label{tab:dcgang}
\begin{center}
\begin{tabular}{@{}rccc@{}}
\toprule
\textbf{Index} & \textbf{Layer} & \textbf{Output Size}  \\ \midrule
$1$ \ \ \ \ & ConvTBlock & $n\times512\times4\times4$\\
$2$ \ \ \ \ & ConvTBlock & $n\times256\times8\times8$\\
$3$ \ \ \ \ & ConvTBlock & $n\times128\times16\times16$\\
$4$ \ \ \ \ & ConvTBlock & $n\times64\times32\times32$\\
$5$ \ \ \ \ & ConvTranpose$4\times4$ & $n\times1\times64\times64$\\
$6$ \ \ \ \ & Tanh &$n\times1\times64\times64$\\
\bottomrule
\end{tabular}

\end{center}
\end{table}
% \FloatBarrier

% \FloatBarrier
\begin{table}[htb!]
\caption{Architecture for DCGAN Discriminator.}
\label{tab:dcgand}
\begin{center}
\begin{tabular}{@{}rccc@{}}
\toprule
\textbf{Index} & \textbf{Layer} & \textbf{Output Size}  \\ \midrule
$1$ \ \ \ \ & ConvBlock & $n\times64\times32\times32$\\
$2$ \ \ \ \ & ConvBlock & $n\times128\times16\times16$\\
$3$ \ \ \ \ & ConvBlock & $n\times256\times8\times8$\\
$4$ \ \ \ \ & ConvBlock & $n\times512\times4\times4$\\
$5$ \ \ \ \ & Conv$4\times4$ & $n\times64\times1\times1$\\
$6$ \ \ \ \ &Reshape &$n\times64$\\
$7$ \ \ \ \ & Linear &$n\times1$\\
$8$ \ \ \ \ & Sigmoid &$n\times1$\\
\bottomrule
\end{tabular}
\end{center}
\end{table}

\begin{table}[htb!]
\caption{Architecture for FID Autoencoder}
\label{tab:extremeae}
\begin{center}
\begin{tabular}{@{}rccc@{}}
\toprule
\textbf{Index} & \textbf{Layer} & \textbf{Output Size}  \\ \midrule
$1$ \ \ \ \ & Linear & $n\times128$\\
$2$ \ \ \ \ & ReLU & $n\times128$\\
$3$ \ \ \ \ & Dropout(0.5) & $n\times128$\\
$4$ \ \ \ \ & Linear & $n\times4096$\\
\bottomrule
\end{tabular}
\end{center}
\end{table}

\section{Implementation Details}
\label{app:3}
The following settings were common to both DCGAN and ExGAN. All convolutional layer weights were initialized from $\mathcal{N}(0, 0.02)$. We sample the noise, or latent inputs, from a standard normal distribution instead of uniform distribution with the latent dimension = $20$. Alpha for LeakyReLU was set to $0.2$. Adam optimizer was used with parameters, Learning rate for $G = 0.0002$, $D = 0.0001$, and betas = ($0.5$, $0.999$). Noisy labels were used, i.e. the Real and Fake labels used for training had values in [$0.7$, $1.2$] and [$0, 0.3$] instead of $1$ and $0$ respectively \cite{noisy}. The Real and Fake labels were flipped with a probability of $0.05$. Gradient clipping was employed restricting the gradients of $G$ and $D$ to be in [-20, 20]. Noise was added to the input of the $D$ starting from $1e-5$ and linearly decreased to $0$. Batch Size was $256$.\\
{\bf Distribution Shifting}: Unless stated otherwise, $c$ was set to 0.75, $k$ was set to 10. For the initial iteration, where the network is trained on all data, the learning rates for $G$ and $D$ were set to $0.0002$ and $0.0001$ respectively and the network was trained for $500$ epochs. For subsequent iterations, learning rates for $G$ and $D$ were lowered to $0.00002$ and $0.00001$ respectively and the network was trained for 100 epochs. \\
FID Autoencoder: The Autoencoder was optimized using Adam with learning rate $0.001$, trained for $50$ epochs with standard L1 Loss. To ensure a fair comparison, we only compare the most extreme samples from DCGAN with ExGAN. Specifically, if ExGAN generates $n$ samples where the extremeness probabilities are sampled uniformly from $(0, \tau]$, then we generate $\lceil\frac{n}{\tau}\rceil$ samples from DCGAN and retain the most extreme $n$ samples for comparison. 

\section{Background}
\label{app:4}
\subsection{GAN and DCGAN:}
Generative Adversarial Network (GAN) is a framework to train deep generative models. The training is done using a minimax game, where a generator $G$ producing synthetic samples plays against a discriminator $D$ that attempts to discriminate between real data and samples created by $G$. The goal of the generator is to learn a distribution $P_G$ which matches the data distribution $P_{data}$. Instead of explicitly estimating $P_{G}$, $G$ learns to transform noise variables $z \sim P_{noise}$, where $P_{noise}$ is the distribution of noise, into synthetic samples $ x \sim G(z)$. The discriminator $D$ outputs $D(x)$ representing the probability of a sample $x$ coming from the true data distribution. In practice, both $G(z;\theta_g)$ and $D(x;\theta_d)$ are parameterized by neural networks. $G$ and $D$ are simultaneously trained by using the minimax game objective $V_{GAN}(D,G)$:
\begin{align*}
   \min _{G} \max _{D} V_{GAN}(D, G)=\mathbb{E}_{x \sim P_{data}}[\log D(x)] \\
   + \ \mathbb{E}_{z \sim P_{n o i s e}}[\log (1-D(G(z)))]
\end{align*}
The stability in training and the effectiveness in learning unsupervised image representations are some of the reasons that make Deep Convolutional GAN, or DCGAN, \cite{radford2016unsupervised} one of the most popular and successful network design for GAN, especially when dealing with image data. The DCGAN model uses strided convolutions in the discriminator and fractional strided convolutions in the generator along with a bunch of tricks to stabilize training. 
\subsection{CGAN and CDCGAN:}
CGAN extends GANs to conditional models by adding auxiliary information, or conditionals, to both the generator and discriminator. It is done by feeding the conditional, $y$, as an additional input layer. The modified objective is given by  
\begin{align*}
    \min _{G} \max _{D} V(D, G)=\mathbb{E}_{{x} \sim P_{data}}[\log D({x} | {y})] \\
   + \ \mathbb{E}_{{z} \sim P_{noise}}[\log (1-D(G({z} | {y})))] 
\end{align*}
The implementation of CGAN consists of linear or fully connected layers. cDCGAN improves on CGAN by using the DCGAN architecture along with the additional conditional input. The use of convolutional layers generates samples with much better image quality compared to CGAN.

\subsection{Extreme Value Theory:}

\begin{theorem} \cite{theorem2}\cite{theorem1}. For a large class of distributions a function $\sigma(u)$ can be found such that
\begin{equation}
\lim _{u \rightarrow \bar{x}} \sup _{0 \leq x<\bar{x}-u}\left|F_{u}(x)-G_{\sigma(u),\xi}\right|=0
\end{equation}
where $\bar{x}$ is the rightmost point of the distribution, $u$ is a threshold, and $F_{u}$ is the \emph{excess distribution function}, i.e. $F_{u}(x)=P(X-u \leq x | X>u)$.
\end{theorem}

\section{Ablation Results}
\label{app:5}
To evaluate the advantage of distribution shifting, we construct a model with an architecture similar to ExGAN, but trained over all images in the dataset, i.e. no Distribution Shifting has been applied. This model is then evaluated in the same manner as described in the paper.

Without distribution shifting, the reconstruction loss remains almost the same as ExGAN ($0.0166$ compared to $0.0172$). However, we observe that the FID score increases significantly ($0.0493 \pm 0.0097$ compared to $0.0236 \pm 0.0037$), showing the need for distribution shifting.

\end{document}